\documentclass{article}

  \usepackage[dblblindworkshop, final]{neurips_2026}

\workshoptitle{The 6th Workshop on Mathematical Reasoning and AI}

\usepackage[utf8]{inputenc} %
\usepackage[T1]{fontenc}    %
\usepackage{hyperref}       %
\usepackage{url}            %
\usepackage{booktabs}
\usepackage{array}       %
\usepackage{amsfonts}       %
\usepackage{nicefrac}       %
\usepackage{microtype}      %
\usepackage{xcolor}         %

\usepackage{csquotes}
\usepackage{zref-clever}
\usepackage{amsmath,amssymb,amsthm}
\usepackage{graphicx}
\usepackage{tikz}
\usetikzlibrary{arrows.meta,positioning,fit,backgrounds,calc}

\usepackage{listings}

\lstdefinestyle{prompt}{
  basicstyle=\ttfamily\footnotesize,
  breakatwhitespace=true,
  breaklines=true,
  columns=fullflexible,
  frame=single,
  framexleftmargin=1.5em,
  keepspaces=true,
  numberblanklines=true,
  numbers=left,
  numbersep=8pt,
  numberstyle=\tiny\color{gray},
  showstringspaces=false,
  stepnumber=1,
  upquote=true,
  xleftmargin=2em
}

\lstdefinestyle{lean}{
  basicstyle=\ttfamily\scriptsize,
  breaklines=true,
  breakatwhitespace=true,
  columns=fullflexible,
  keepspaces=true,
  frame=single,
  showstringspaces=false,
  upquote=true,
  escapeinside={(*@}{@*)}
}

\usepackage{pifont}
\zcsetup{cap}
\newcommand{\PaperVerified}{\textcolor{green!50!black}{\ding{51}}}
\newcommand{\PaperUnverified}{\textcolor{red!75!black}{\ding{55}}}
\newcommand{\PaperMiniSuccesses}{7}
\newcommand{\PaperAxSuccesses}{4}
\newcommand{\PaperMiniCostPerProof}{\$2.09}
\newcommand{\PaperAxCostPerProof}{\$5.24}

\newcommand{\PaperAstraSuccesses}{12}
\newcommand{\PaperAstraCharge}{\$15.37}
\newcommand{\PaperAstraCostPerProof}{\$1.28}
\newcommand{\PaperAstraCalls}{106}
\newcommand{\PaperAstraAFiveCharge}{\$2.69}
\newcommand{\PaperAstraBSixCharge}{\$1.62}
\newcommand{\PaperFableSuccesses}{11}
\newcommand{\PaperFableCharge}{\$120.29}
\newcommand{\PaperFableCostPerProof}{\$10.94}
\newcommand{\PaperFableAFiveCharge}{\$50.10}
\newcommand{\PaperFableAFiveCalls}{35}
\newcommand{\PaperFableAFiveMinutes}{40.16}
\newcommand{\PaperMiniBashRequests}{1,088}
\newcommand{\PaperMiniCatLeanRequests}{602}
\newcommand{\PaperMiniSearchRequests}{73}
\newcommand{\PaperMiniCalls}{1,091}
\newcommand{\PaperAxCalls}{724}
\newcommand{\PaperMiniReasoningMillion}{0.70}
\newcommand{\PaperAxReasoningMillion}{3.64}
\newcommand{\PaperMiniCachedInputPercent}{91.5\%}
\newcommand{\PaperAxCachedInputPercent}{8.7\%}
\newcommand{\PaperAxAuxiliarySpendPercent}{11.1\%}
\newcommand{\PaperAxBudgetIncomplete}{6}
\newcommand{\PaperLowBudgetTable}{%
\begin{tabular}{@{}lcccc@{}}
\toprule
Harness & \$0.50 & \$1.00 & \$2.00 & Charge \\
\midrule
mini-Lean-agent & 0/12 & 4/12 & 7/12 & \$14.64 \\
AxProverBase & 0/12 & 0/12 & 4/12 & \$20.97 \\
\bottomrule
\end{tabular}
}
\newcommand{\PaperPerProblemTable}{%
\begin{tabular}{@{}lcccc@{}}
\toprule
Problem & mini Verified & mini Charge & Ax Verified & Ax Charge \\
\midrule
A1 & \PaperVerified{} & \$0.518454 & \PaperVerified{} & \$2.001509 \\
A2 & \PaperVerified{} & \$0.658883 & \PaperUnverified{} & \$0.141243 \\
A3 & \PaperVerified{} & \$0.677093 & \PaperVerified{} & \$1.758844 \\
A4 & \PaperVerified{} & \$1.064088 & \PaperUnverified{} & \$2.014457 \\
A5 & \PaperUnverified{} & \$2.042195 & \PaperUnverified{} & \$2.000157 \\
A6 & \PaperVerified{} & \$1.192095 & \PaperUnverified{} & \$1.949676 \\
B1 & \PaperUnverified{} & \$2.017090 & \PaperUnverified{} & \$2.006616 \\
B2 & \PaperUnverified{} & \$2.012949 & \PaperUnverified{} & \$2.063717 \\
B3 & \PaperVerified{} & \$0.657564 & \PaperUnverified{} & \$2.115382 \\
B4 & \PaperVerified{} & \$1.622776 & \PaperVerified{} & \$1.105187 \\
B5 & \PaperUnverified{} & \$0.149430 & \PaperVerified{} & \$1.814421 \\
B6 & \PaperUnverified{} & \$2.023811 & \PaperUnverified{} & \$2.003222 \\
\bottomrule
\end{tabular}
}
\newcommand{\PaperExtendedTable}{%
\begin{tabular}{@{}lccccc@{}}
\toprule
Problem & Harness & Verified & Charge & Calls & Hours \\
\midrule
A5 & mini-Lean-agent & \PaperUnverified{} & \$10.773974 & 300 & 1.954 \\
A5 & AxProverBase & \PaperUnverified{} & \$8.485967 & 300 & 1.653 \\
B6 & mini-Lean-agent & \PaperUnverified{} & \$3.268421 & 137 & 0.816 \\
B6 & AxProverBase & \PaperUnverified{} & \$6.609995 & 173 & 1.254 \\
\bottomrule
\end{tabular}
}
\newcommand{\PaperHighBudgetTable}{%
\begin{tabular}{@{}lcrcc@{}}
\toprule
Model & Verified & Charge & Calls & Minutes \\
\midrule
Astra & 12/12 & \$15.37 & 106 & 51.3 \\
Fable 5.1 & 11/12 & \$120.29 & 133 & 99.9 \\
\bottomrule
\end{tabular}
}
\newcommand{\PaperHighBudgetPerProblemTable}{%
\begin{tabular}{@{}lcccccccc@{}}
\toprule
 & \multicolumn{4}{c}{Astra} & \multicolumn{4}{c}{Fable 5.1} \\
\cmidrule(lr){2-5} \cmidrule(lr){6-9}
Problem & Verified & Charge & Calls & Minutes & Verified & Charge & Calls & Minutes \\
\midrule
A1 & \PaperVerified{} & \$0.61 & 5 & 3.92 & \PaperVerified{} & \$5.09 & 13 & 9.81 \\
A2 & \PaperVerified{} & \$1.58 & 9 & 6.16 & \PaperVerified{} & \$5.98 & 9 & 12.78 \\
A3 & \PaperVerified{} & \$0.96 & 9 & 8.01 & \PaperVerified{} & \$3.16 & 5 & 10.33 \\
A4 & \PaperVerified{} & \$0.98 & 7 & 6.83 & \PaperVerified{} & \$4.94 & 8 & 11.91 \\
A5 & \PaperVerified{} & \$2.69 & 23 & 17.43 & \PaperUnverified{} & \$50.10 & 35 & 40.16 \\
A6 & \PaperVerified{} & \$0.79 & 5 & 5.06 & \PaperVerified{} & \$7.53 & 9 & 14.54 \\
B1 & \PaperVerified{} & \$1.13 & 8 & 7.80 & \PaperVerified{} & \$10.32 & 11 & 19.12 \\
B2 & \PaperVerified{} & \$2.06 & 12 & 11.35 & \PaperVerified{} & \$6.42 & 10 & 15.86 \\
B3 & \PaperVerified{} & \$0.62 & 5 & 3.37 & \PaperVerified{} & \$1.56 & 6 & 5.33 \\
B4 & \PaperVerified{} & \$1.03 & 6 & 6.78 & \PaperVerified{} & \$10.54 & 7 & 21.72 \\
B5 & \PaperVerified{} & \$1.31 & 8 & 8.04 & \PaperVerified{} & \$5.01 & 6 & 12.65 \\
B6 & \PaperVerified{} & \$1.62 & 9 & 10.27 & \PaperVerified{} & \$9.64 & 14 & 17.09 \\
\bottomrule
\end{tabular}
}
\newcommand{\PaperHighBudgetCheckpointsTable}{%
\begin{tabular}{@{}lccccc@{}}
\toprule
Model & \$2 & \$5 & \$10 & \$20 & \$50 \\
\midrule
Astra & 10/12 & 12/12 & 12/12 & 12/12 & 12/12 \\
Fable 5.1 & 1/12 & 3/12 & 9/12 & 11/12 & 11/12 \\
\bottomrule
\end{tabular}
}
\newcommand{\PaperChargesTable}{%
\begin{tabular}{@{}lc@{}}
\toprule
Model & Charge \\
\midrule
Gemini 3.8 Flash & \$24.612816 \\
Gemini 3.8 Flash (AxProverBase) & \$32.067014 \\
Opus 4.5 & \$61.596995 \\
Astra & \$15.367271 \\
Fable 5.1 & \$120.294990 \\
\midrule
Total & \$253.939087 \\
\bottomrule
\end{tabular}
}
\newcommand{\PaperToolUseTable}{%
\begin{tabular}{@{}lccccccc@{}}
\toprule
Problem & Bash & Search & \texttt{cat} & \texttt{sed} & Python & Lean & Find/list \\
\midrule
A1 & 63 & 0 & 59 & 0 & 3 & 57 & 0 \\
A2 & 75 & 4 & 62 & 1 & 2 & 61 & 1 \\
A3 & 45 & 1 & 38 & 3 & 0 & 37 & 1 \\
A4 & 79 & 0 & 53 & 1 & 22 & 54 & 1 \\
A5 & 141 & 16 & 68 & 2 & 40 & 66 & 11 \\
A6 & 86 & 5 & 31 & 0 & 49 & 29 & 1 \\
B1 & 142 & 5 & 115 & 9 & 55 & 67 & 2 \\
B2 & 181 & 33 & 132 & 4 & 0 & 132 & 7 \\
B3 & 68 & 6 & 40 & 1 & 15 & 41 & 1 \\
B4 & 92 & 1 & 47 & 1 & 38 & 46 & 3 \\
B5 & 17 & 0 & 5 & 0 & 8 & 4 & 3 \\
B6 & 99 & 2 & 24 & 0 & 54 & 23 & 13 \\
Total & 1,088 & 73 & 674 & 22 & 286 & 617 & 44 \\
\bottomrule
\end{tabular}
}
\newcommand{\PaperUsageTable}{%
\begin{tabular}{@{}lcc@{}}
\toprule
Quantity & mini-Lean-agent & AxProverBase \\
\midrule
Model calls & 1,091 & 724 \\
Uncached input tokens (M) & 6.47 & 8.11 \\
Cached input tokens (M) & 69.69 & 0.78 \\
Visible output tokens (M) & 0.52 & 0.32 \\
Reasoning tokens (M) & 0.70 & 3.64 \\
Cached share of input & 91.5\% & 8.7\% \\
Mean charge per model call & \$0.0134 & \$0.0290 \\
\bottomrule
\end{tabular}
}

\title{Do Theorem Provers Need Theorem-Proving Agents?\\ mini-Lean-agent: A Coding-Agent Baseline for Lean}

\author{%
  Tomas Ortega\\
  Princeton Language and Intelligence\\
  Princeton University\\
  Princeton, NJ 08540 \\
  \texttt{tomas.ortega@princeton.edu} \\
  \And
  Simon Park\\
  Princeton Language and Intelligence\\
  Princeton University\\
  Princeton, NJ 08540 \\
  \texttt{juhyunp@princeton.edu} \\
  \And
  Ayush Khaitan\\
  Princeton Language and Intelligence\\
  Princeton University\\
  Princeton, NJ 08540 \\
  \texttt{ayush.khaitan@princeton.edu} \\
  \And
  Liam Fowl\\
  Princeton Language and Intelligence\\
  Princeton University\\
  Princeton, NJ 08540 \\
  \texttt{lf2728@princeton.edu} \\
  \And
  Alex Kontorovich\\
  Department of Mathematics\\
  Rutgers University\\
  Piscataway, NJ 08854\\
  \texttt{alex.kontorovich@rutgers.edu}
  \And
  Sanjeev Arora\\
  Princeton Language and Intelligence\\
  Princeton University\\
  Princeton, NJ 08540 \\
  \texttt{arora@princeton.edu} \\
}

\begin{document}

\maketitle

\begin{abstract}
  General-purpose coding agents use standard shell and file interactions, yet formal theorem-proving agents often rely on specialized, ad-hoc tooling.
  This makes it difficult to determine which components of theorem-proving agents provide benchmark gains: an improved base model, a specialized harness, a specific tool, or a combination of any of them.
  We introduce mini-Lean-agent, a deliberately minimal agent for Lean tasks whose only model-visible tool is bash.
  The model must use bash to inspect and edit the repository, search local libraries, invoke Lean, and manage its own intermediate work, while a separate trusted evaluator verifies the final artifact.
  We compare mini-Lean-agent against AxProverBase and Numina-Lean-Agent on the Putnam 2025 benchmark.
  Experiments with Gemini 3.8 Flash as the base model show that mini-Lean-agent solves \PaperMiniSuccesses{}/12 problems at \$2 per problem limit, compared with AxProverBase's \PaperAxSuccesses{}/12.
  With Astra as the base model, mini-Lean-agent solves 12/12 problems at an average cost of \$1.28 per problem.
  This suggests that a generic coding agent with a minimal harness can already achieve strong performance on Lean benchmarks.
  Our results motivate using mini-Lean-agent as a standard baseline when evaluating theorem-proving infrastructure.
\end{abstract}

\section{Introduction}
The past few months have seen a flurry of mathematics results solved by Large Language Models (LLMs), both in natural language and in formal language (most notably Lean~\citep{Moura_Ullrich_2021}).
In practice, mathematicians and engineers who use LLMs for Lean rely on general-purpose coding agents (such as Claude Code or Codex) rather than specialized prover-specific harnesses.
However, the AI-for-math literature proposes a variety of specialized machinery to construct agents for Lean tasks: special-purpose harnesses~\citep{lample_hypertree_2022,thakur_-context_2024,pegden_autoformalization_2026,li_merlean-prover_2026}, specialized tools~\citep{requena_minimal_2026,hu_minictx_2025}, or a combination of both~\citep{yang_leandojo_2023,lin_goedel-prover_2025,lin_goedel-prover-v2_2025,chung_goedel-architect_2026,aniva_pantograph_2025,liu_numina-lean-agent_2026,dekoninck_beyond_2026,tsoukalas_advancing_2026,song_lean_2025}.
These works report substantial performance improvements over variations of their predecessors, but it is hard to attribute the gains to an improved base model, the harness, or the specialized tools considered in isolation.

In software engineering, this problem has been addressed by the development of general-purpose coding agents with a minimal prompt (less than 100 lines) that operate through a shell and filesystem, most notably mini-SWE-agent by~\citet{yang_swe-agent_2024}.
We then ask: \textbf{Should we use a generic coding agent for evaluating Lean-specific agent machinery?}
Rather than positioning our work on the spectrum of few versus many tools, we position it as \emph{generic computer access versus Lean-specific tools}.
First, this establishes a baseline for Lean-agent research that is more comparable to the workflow of current practitioners.
Second, it helps isolate the contribution of the base model from the contribution of specialized tools or harnesses.

To study this, we introduce mini-Lean-agent, a deliberately minimal baseline whose only model-visible tool is \texttt{bash}.
Note that a generic coding agent with a shell can naturally replicate what specialized tooling hard-codes: searching Mathlib with \texttt{rg} or \texttt{grep}, running tests and interpreting compiler errors with \texttt{lake env lean}, creating temporary scratchpads (e.g., \texttt{Scratch.lean}), and transferring successful proofs to the target file.
Because the model can modify its own environment, the output artifact is treated as untrusted: a separate, trusted evaluator validates the final proof in a clean environment using Comparator~\citep{noauthor_leanprovercomparator_2026} to catch build failures, undeclared axioms, or statement mutations.

Using this filesystem-native baseline, we investigate two primary research questions:
\begin{itemize}
  \item \textbf{RQ1 (Parity and Cost):} Using the same model and inference budget, how much solve rate and cost-efficiency does mini-Lean-agent recover relative to specialized theorem-proving baselines?
  \item \textbf{RQ2 (Minimal-Scaffold Capability):}  How far can stronger frontier models go on Lean benchmarks using only a minimal coding agent scaffold?
\end{itemize}

Our code is available at \url{https://github.com/TomasOrtega/mini-lean-agent}.

After mini-Lean-agent was made public, a concurrent project, \texttt{h1}~\cite{moalla2026h1}, was released.
Like our work, \texttt{h1} argues that minimal agent harnesses achieve strong theorem-proving performance with current models.
However, \texttt{h1} requires specialized tools for Lean library search, whereas mini-Lean-agent relies exclusively on \texttt{bash}.

\section{System Design}
To separate the contribution of the model from that of the harness, we keep the agent interface to a shell and handle proof acceptance independently.

\paragraph{Filesystem-native agent.}
mini-Lean-agent is built on mini-SWE-agent, whose only model-visible tool is \texttt{bash}.
Given a file containing the target theorem and a pinned Lean environment, the model can inspect and edit files, search local libraries, invoke Lean, and manage its own intermediate work.
\zcref{fig:agent-interfaces} contrasts this generic computer access with the Lean-specific interface of AxProverBase.
For example, mini-Lean-agent can create a \texttt{Scratch.lean} file to test intermediate results before incorporating successful proofs into the target file.
AxProverBase instead provides two custom designed search tools, one for a fuzzy search on vectorized embeddings of Lean statements \citep{gao-etal-2024-semantic-search} and another for a web search of proof strategies. It also separates the workflow into multiple modules, including a proposer (which writes Lean proofs), a compiler (which compiles the Lean proof), and a memory module (where the compiler logs the previous errors).
The main distinction is the kind of operations the model can perform and compose during proof development, not just the number of tools it has available.

\begin{figure}[htbp]
  \centering
  \usetikzlibrary{arrows.meta,calc,shapes.geometric,shapes.symbols}
  \resizebox{\linewidth}{!}{%
    \begingroup
\definecolor{mlaInk}{HTML}{273746}
\definecolor{mlaBlue}{HTML}{376A87}
\definecolor{mlaPlum}{HTML}{785C7D}
\definecolor{mlaRule}{HTML}{A3ACB4}
\definecolor{mlaPaper}{HTML}{F5F7F8}
\begin{tikzpicture}[
  x=1cm, y=1cm,
  font=\fontsize{8.5}{10}\selectfont,
  text=mlaInk,
  line cap=round, line join=round,
  every node/.style={inner sep=0pt, outer sep=0pt, align=center},
  edge/.style={draw=mlaInk!80, line width=.65pt},
  exchange/.style={edge, {Latex[length=1.35mm,width=1.0mm]}-{Latex[length=1.35mm,width=1.0mm]}},
  submit/.style={edge, -{Latex[length=1.35mm,width=1.0mm]}},
  enclosure/.style={draw=mlaRule, line width=.6pt,
    dash pattern=on 2.3pt off 2.3pt, rounded corners=4pt},
  module/.style={draw=mlaPlum!65, fill=mlaPlum!5,
    rounded corners=2pt, line width=.55pt,
    minimum width=1.74cm, minimum height=.43cm,
    font=\fontsize{8}{9}\selectfont},
  model/.style={draw=mlaInk!75, fill=white, line width=.7pt,
    rounded corners=3pt, minimum width=2.35cm, minimum height=.74cm},
  tool/.style={line width=.75pt, rounded corners=3pt,
    minimum height=.9cm},
  small/.style={font=\fontsize{7.3}{8.5}\selectfont},
  title/.style={font=\bfseries\fontsize{10.5}{12}\selectfont},
  subtitle/.style={font=\fontsize{8}{9}\selectfont, text=mlaInk!80},
  pics/network/.style={code={
    \foreach \ya in {-.17,0,.17}{
      \foreach \yb in {-.13,.13}{
        \draw[draw=mlaInk!50,line width=.35pt] (-.23,\ya)--(0,\yb);
      }
    }
    \foreach \ya in {-.13,.13}{
      \draw[draw=mlaInk!50,line width=.35pt] (0,\ya)--(.23,0);
    }
    \foreach \y in {-.17,0,.17}{\fill[mlaInk](-.23,\y)circle(.035);}
    \foreach \y in {-.13,.13}{\fill[mlaInk](0,\y)circle(.035);}
    \fill[mlaInk](.23,0)circle(.035);
  }},
  pics/document/.style={code={
    \path[fill=white,draw=mlaInk!75,line width=.6pt]
      (-.24,-.31)--(.24,-.31)--(.24,.15)--(.08,.31)--(-.24,.31)--cycle;
    \draw[draw=mlaInk!65,line width=.5pt](.08,.31)--(.08,.15)--(.24,.15);
    \foreach \y in {-.13,-.02,.09}{
      \draw[draw=mlaInk!55,line width=.5pt](-.14,\y)--(.11,\y);
    }
  }},
  pics/books/.style={code={
    \foreach \y in {-.20,0,.20}{
      \path[fill=white,draw=mlaInk!75,line width=.6pt,rounded corners=.8pt]
        (-.32,\y-.055) rectangle (.30,\y+.075);
      \draw[draw=mlaInk!50,line width=.45pt](-.20,\y-.045)--(-.20,\y+.065);
    }
  }},
  pics/compiler/.style={code={
    \path[draw=mlaInk!75,fill=white,line width=.65pt,rounded corners=1.5pt]
      (-.55,-.27)rectangle(.55,.27);
    \node[font=\bfseries\fontsize{8}{9}\selectfont]at(0,0){Lean};
  }},
  pics/magnifier/.style={code={
    \draw[draw=mlaPlum,line width=.8pt](-.045,.035)circle(.115);
    \draw[draw=mlaPlum,line width=.95pt](.036,-.046)--(.155,-.165);
  }},
  pics/shield/.style={code={
    \path[draw=mlaInk!75,fill=white,line width=.65pt]
      (-.22,.22)--(0,.28)--(.22,.22)--(.20,-.055)
      ..controls(.15,-.20)and(.06,-.24)..(0,-.27)
      ..controls(-.06,-.24)and(-.15,-.20)..(-.20,-.055)--cycle;
    \draw[draw=mlaInk,line width=.95pt](-.105,.005)--(-.02,-.08)--(.12,.095);
  }}
]

\node[title,text=mlaBlue] at (3.4,8.48) {(a) mini-Lean-agent};
\node[subtitle] at (3.4,8.10) {Generic computer access};
\node[title,text=mlaPlum] at (10.6,8.48) {(b) AxProverBase};
\node[subtitle] at (10.6,8.10) {Harness-defined proof interactions};

\draw[enclosure] (.10,4.59) rectangle (6.70,7.76);
\draw[enclosure] (7.30,4.59) rectangle (13.90,7.76);
\node[small,fill=white,inner xsep=5pt] at (3.4,7.76) {Minimal agent loop};
\node[small,fill=white,inner xsep=5pt] at (10.6,7.76) {Configured agent harness};

\node[subtitle] at (3.4,7.19) {Workflow composed by the model};

\node[module] (refine) at (8.55,7.19) {Refinement};
\node[module] (review) at (10.60,7.19) {Review};
\node[cylinder,shape border rotate=90,aspect=.28,
  draw=mlaPlum!65,fill=mlaPlum!5,line width=.55pt,
  minimum width=1.70cm,minimum height=.46cm,
  font=\fontsize{8}{9}\selectfont]
  (memory) at (12.65,7.19) {Memory};

\node[model] (mini-model) at (3.4,6.43) {};
\pic at (2.75,6.43) {network};
\node[font=\bfseries\fontsize{10}{11}\selectfont] at (3.70,6.43) {LLM};
\node[model] (ax-model) at (10.6,6.43) {};
\pic at (9.95,6.43) {network};
\node[font=\bfseries\fontsize{10}{11}\selectfont] at (10.90,6.43) {LLM};

\draw[draw=mlaPlum!50,line width=.45pt,densely dotted]
  (refine.south)--(ax-model.north west);
\draw[draw=mlaPlum!50,line width=.45pt,densely dotted]
  (review.south)--(ax-model.north);
\draw[draw=mlaPlum!50,line width=.45pt,densely dotted]
  (memory.south)--(ax-model.north east);

\node[tool,draw=mlaBlue,fill=mlaBlue!4,minimum width=3.35cm]
  (shell) at (3.4,5.19) {};
\draw[draw=mlaBlue!45,line width=.45pt] (1.73,5.44)--(5.07,5.44);
\foreach \x in {1.91,2.06,2.21}{\fill[mlaBlue!60](\x,5.57)circle(.026);}
\node[font=\ttfamily\bfseries\fontsize{11}{11}\selectfont,text=mlaBlue]
  at (2.10,5.09) {>\_};
\node[font=\ttfamily\bfseries\fontsize{10}{11}\selectfont,text=mlaBlue]
  at (3.70,5.18) {bash};
\node[small] at (3.70,4.92) {only exposed tool};

\node[tool,draw=mlaPlum,fill=mlaPlum!4,minimum width=3.25cm]
  (feedback) at (9.33,5.19) {};
\node[font=\bfseries\fontsize{8.5}{10}\selectfont,text=mlaPlum]
  at (9.33,5.33) {Proof interaction};
\node[font=\fontsize{7}{8.5}\selectfont] at (9.33,5.04) {structured Lean feedback};
\node[tool,draw=mlaPlum,fill=mlaPlum!4,minimum width=2.45cm]
  (search) at (12.40,5.19) {};
\pic at (11.60,5.27) {magnifier};
\node[anchor=west,font=\bfseries\fontsize{7.5}{9}\selectfont,text=mlaPlum]
  at (11.82,5.29) {LeanSearch};
\node[small] at (12.40,5.00) {search tool};

\draw[exchange] (mini-model.south)--(shell.north);
\draw[exchange] (ax-model.south)--(10.60,5.85)-|(feedback.north);
\draw[exchange] (ax-model.east)-|(search.north);

\node[cloud,cloud puffs=11,cloud puff arc=110,aspect=4,
  draw=mlaRule,fill=white,line width=.55pt,
  minimum width=2.3cm,minimum height=.44cm,
  font=\fontsize{7.4}{8.5}\selectfont]
  (service) at (12.40,3.98) {Search service};
\draw[edge, {Latex[length=1mm,width=.8mm]}-{Latex[length=1mm,width=.8mm]}]
  (search.south)--(service.north);

\path[draw=mlaBlue!65,fill=mlaBlue!3,line width=.65pt,rounded corners=2pt]
  (.25,1.60)--(.25,3.67)--(4.98,3.67)--(5.25,3.40)
  --(6.55,3.40)--(6.55,1.60)--cycle;
\node[font=\bfseries\fontsize{8.3}{9.5}\selectfont,
  anchor=west] at (.45,3.49) {Model-controlled workspace};

\pic[scale=.74] at (.88,2.80) {document};
\pic[scale=.74] at (.88,2.01) {document};
\pic[scale=.80] at (3.96,2.80) {books};
\pic[scale=.76] at (3.96,2.01) {compiler};
\node[anchor=west,font=\fontsize{8.3}{9.4}\selectfont]
  at (1.30,2.92) {Target proof};
\node[anchor=west,small,text=mlaInk!80]
  at (1.30,2.65) {inspect / edit};
\node[anchor=west,font=\bfseries\fontsize{8.3}{9.4}\selectfont,
  text=mlaBlue] at (1.30,2.13) {Scratch files};
\node[anchor=west,small,text=mlaBlue]
  at (1.30,1.86) {create / test};
\node[anchor=west,font=\fontsize{8.3}{9.4}\selectfont]
  at (4.45,2.92) {Mathlib};
\node[anchor=west,small,text=mlaInk!80]
  at (4.45,2.65) {search sources};
\node[anchor=west,font=\fontsize{8.3}{9.4}\selectfont]
  at (4.45,2.13) {Compiler};
\node[anchor=west,small,text=mlaInk!80]
  at (4.45,1.86) {invoke Lean};

\path[draw=mlaBlue,line width=.55pt,fill=white]
  (1.05,2.18) circle (.085);
\draw[draw=mlaBlue,line width=.7pt]
  (1.005,2.18)--(1.095,2.18) (1.05,2.135)--(1.05,2.225);

\path[draw=mlaRule,fill=mlaPaper,line width=.65pt,rounded corners=3pt]
  (7.45,1.60)rectangle(13.75,3.40);
\node[font=\bfseries\fontsize{8.3}{9.5}\selectfont]
  at (10.60,3.12) {Harness-managed proof checking};
\draw[draw=mlaRule!50,line width=.45pt] (7.65,2.91)--(13.55,2.91);
\pic at (9.00,2.39) {document};
\pic at (12.16,2.39) {compiler};
\node[font=\fontsize{8}{9}\selectfont]
  at (9.00,1.86) {Proof attempt};
\node[font=\fontsize{8}{9}\selectfont]
  at (12.16,1.86) {Lean compiler};
\draw[exchange] (9.46,2.39)--(11.48,2.39);
\node[font=\fontsize{7}{8.5}\selectfont,fill=mlaPaper,inner xsep=1pt]
  at (10.42,2.65) {check / feedback};

\draw[exchange] (shell.south)--(3.4,3.67);
\node[small,anchor=west,align=left,fill=white,inner sep=2pt]
  at (3.58,4.13) {file operations\\compiler output};
\draw[exchange] (feedback.south)--(9.33,3.89)--(10.60,3.89)--(10.60,3.40);
\node[small,anchor=east,align=right,fill=white,inner sep=2pt]
  at (9.15,4.13) {proof attempts\\diagnostics};
\end{tikzpicture}
\endgroup%
  }
  \caption{
    \textbf{A comparison between mini-Lean-agent and AxProverBase workflow.}
    mini-Lean-agent is only given \texttt{bash} tools, through which it can directly search for local Mathlib statements and edit files containing the target proof. AxProverBase instead introduces customized search tools and specialized modules in the workflow. 
  }
  \zlabel{fig:agent-interfaces}
\end{figure}
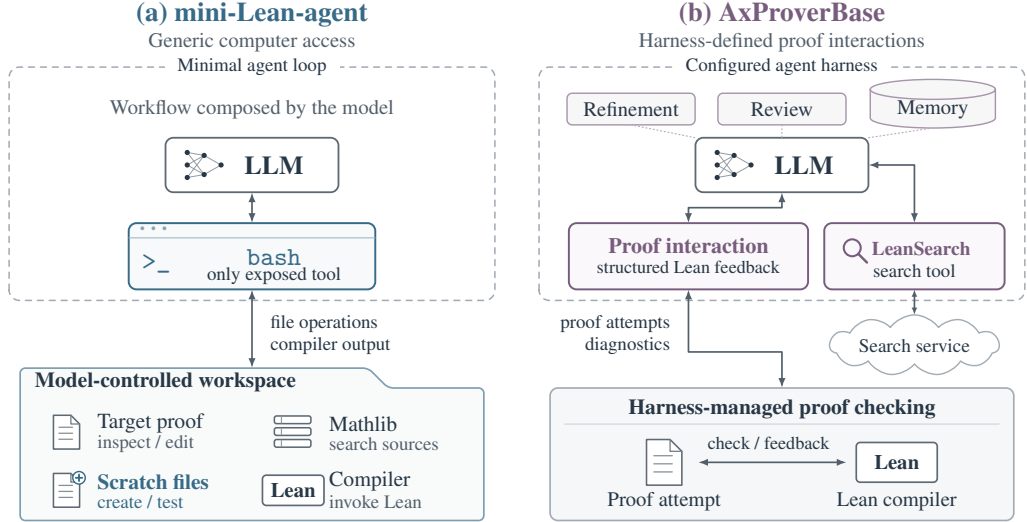

\paragraph{Trusted evaluation.}
Because the agent can edit its workspace, we treat its output file as untrusted.
After each run, a separate evaluator copies only the declared submission artifact into a fresh environment with the original theorem statement and pinned Lean and Mathlib versions.
Comparator~\citep{noauthor_leanprovercomparator_2026} checks the candidate against the original theorem statement.
This allows us to detect and reject undeclared axioms, \texttt{sorry}, \texttt{admit}, build failures, and timeouts.

\section{Experimental Design}
\zlabel{sec:experiment}

We use this procedure to evaluate mini-Lean-agent on the Putnam 2025 benchmark. For RQ1, we hold the model and budget fixed while comparing against AxProverBase, a different harness. For RQ2, we fix mini-Lean-agent and vary the model. We additionally compare against the reported results from Numina-Lean-Agent~\citep{liu_numina-lean-agent_2026}.

\paragraph{Tasks, models, and tools.}
We use the twelve Putnam 2025 challenge files A1--A6 and B1--B6 released with Numina-Lean-Agent, preserving the statement formalizations it attributes to Seed-Prover 1.5~\citep{liu_numina-lean-agent_2026}.
We use Lean 4.28.0-rc1 and Mathlib \texttt{v4.28.0-rc1} throughout.
We run each problem-agent pair in its own Docker container and disable open-web access and search grounding.

\paragraph{Changing the harness, while keeping the model fixed.}
We use Gemini 3.8 Flash with \texttt{high} thinking as the base model for both mini-Lean-agent and AxProverBase.
We report one run per problem and harness.
We stop when the agent finishes or reaches one of the following limits per problem: \$2 model charge, 200 model calls, or 120 minutes of active time. We allow the final request to finish if it began below the model charge limit, and count any resulting proof as long as the call and time limits still hold.
For A5 and B6, which are known to be harder problems~\citep{liu_numina-lean-agent_2026}, we allow up to \$50 model charge, 300 calls, and 24 active hours per problem (continuing from the model output when the limits have been reached at a lower budget). 

\paragraph{Scaling to stronger models, while keeping harness fixed.}
To test how much mini-Lean-agent can achieve with stronger base models, we run it with Astra and Fable 5.1 at \texttt{max} reasoning effort on all twelve problems, with a maximum budget of \$50 model charge, 200 model calls (300 for A5 and B6), and 24 active hours per problem.
We compare these complete systems with Numina's reported performance and resources.

\section{Results}

\subsection{RQ1: A minimal coding agent outperforms a minimal agent with Lean tooling}
With Gemini and a \$2 limit, mini-Lean-agent and AxProverBase verify \PaperMiniSuccesses{}/12 and \PaperAxSuccesses{}/12 runs, respectively (\zcref{tab:controlled}).
The charge per successful run is \PaperMiniCostPerProof{} versus \PaperAxCostPerProof{}.
Note that if the output is truncated at an even lower per-problem budget of \$1, AxProverBase would not have been able to solve any problem, whereas mini-Lean-agent can still solve 4.

\begin{table}[htbp]
  \centering
  \caption{\textbf{Comparison between mini-Lean-agent and AxProverBase with Gemini at \$2 limit.} Each column (except last) shows the number of verified problems out of 12 when the output is truncated at an even smaller per-problem limit. The last column shows the total charge, including that for unsuccessful attempts. This does not include the charge for the higher budget run for A5/B6.}
  \zlabel{tab:controlled}
  \PaperLowBudgetTable
\end{table}

For either harness, running with an extra budget on A5 and B6 does not result in a correct proof. 
For extended details, \zcref{tab:perproblem,tab:long-horizon} give the per-problem results and continuations in the appendix.

The bash traces show how mini-Lean-agent develops its proofs (for details, see \zcref{tab:rq1-tools} in the appendix).
Of \PaperMiniBashRequests{} requests, \PaperMiniCatLeanRequests{} combine writing a file with \texttt{cat} and invoking Lean, and only \PaperMiniSearchRequests{} request \texttt{rg} or \texttt{grep}.
All twelve runs use scratch Lean files.
In A1, the model tests a finite-set lemma with \texttt{exact?} in \texttt{Scratch.lean}, then uses Lean's suggested lemma in the accepted proof.
The excerpts with the key interactions are shown in \zcref{fig:rq1-excerpts}, in the appendix.

A possible explanation for the difference in cost is that mini-Lean-agent is more efficient in its model usage.
AxProverBase makes fewer model calls (\PaperAxCalls{} versus \PaperMiniCalls{}), but generates \PaperAxReasoningMillion{} million reasoning tokens versus \PaperMiniReasoningMillion{} million (see \zcref{tab:rq1-usage} in the appendix for a full breakdown).
Cache reads account for \PaperMiniCachedInputPercent{} of mini-Lean-agent's input tokens, versus \PaperAxCachedInputPercent{} for AxProverBase.
Memory and review calls account for \PaperAxAuxiliarySpendPercent{} of AxProverBase's charge.
All \PaperAxBudgetIncomplete{} unsuccessful AxProverBase runs that exhaust the allowance still leave \texttt{sorry} in their final candidates.
On B3, two of eight proof responses truncate mid-JSON, and the final candidate leaves a modular-arithmetic case unresolved.
For the excerpts, we refer the reader again to \zcref{fig:rq1-excerpts} in the appendix.

\subsection{RQ2: mini-Lean-agent with a strong model is a competitive baseline}
Since Gemini fails on some problems, we naturally explore whether mini-Lean-agent can solve them with a stronger model.
With Astra, we verify \PaperAstraSuccesses{}/12 problems for \PaperAstraCharge{} in total, or \PaperAstraCostPerProof{} per verified proof (\zcref{tab:high-budget}).
Astra solves every problem for under \$3, including A5 for \PaperAstraAFiveCharge{} and B6 for \PaperAstraBSixCharge{}, and it uses \PaperAstraCalls{} model calls across the suite.
With Fable 5.1, we verify \PaperFableSuccesses{}/12 problems for \PaperFableCharge{}, or \PaperFableCostPerProof{} per verified proof including the unsuccessful attempt.
Fable leaves only A5 unsolved, reaching \PaperFableAFiveCharge{} after \PaperFableAFiveCalls{} calls and \PaperFableAFiveMinutes{} active minutes while developing an incomplete proof.
\zcref{tab:high-budget-per-problem,tab:high-budget-checkpoints} give the per-problem results and cost checkpoints.

\begin{table}[htbp]
  \centering
  \caption{\textbf{Performance of mini-Lean-agent with Astra and Fable at \$50 limit.}
  The total charge includes the charge for unsuccessful attempts.}
  \zlabel{tab:high-budget}
  \PaperHighBudgetTable
\end{table}

The 12/12 result using mini-Lean-agent and Astra is on par with the result of Numina-Lean-Agent~\citep{liu_numina-lean-agent_2026}, the first open-source system (to the best of our knowledge) that achieved 12/12 on Putnam 2025. While the exact accounting of the cost is not offered, the authors report that Numina-Lean-Agent uses an approximate budget of \$50 per task~\citep[Sec.~3.1]{liu_numina-lean-agent_2026}. On the two challenging problems, they report using \$1,000 and 2,040 minutes for A5, and \$300 and 797 minutes for B6. Instead, our mini-Lean-agent with Astra is able to solve all 12 problems with only \$15.37 in total \footnote{Note that Numina-Lean-Agent used Opus 4.5. Because the models and budgets differ, this comparison shows what the complete systems achieve but does not isolate the effect of their harnesses. See \zcref{app:results} for details on a preliminary attempt to run mini-Lean-agent with Opus 4.5.}.

It is worth noting that we cannot rule out dataset contamination, since the versions of Astra, Fable, and Gemini that we use may have been trained on the answers of Putnam 2025.

\section{Conclusion}
We present a minimal coding-agent baseline for Lean, mini-Lean-agent, whose only model-visible tool is \texttt{bash}.
Our experiments show that mini-Lean-agent achieves better performance than AxProverBase on the Putnam 2025 benchmark at a lower cost per verified proof, and that it can solve all twelve problems with Astra at an average cost of \$1.28 per problem.
This suggests that a generic coding agent with a minimal harness can serve as a strong baseline for Lean benchmarks.
An important question for future work is how the value of specialized theorem-proving scaffolding changes as base models become more capable.

\begin{ack}
    The authors acknowledge funding from the DARPA expMath program. We thank Dr. Patrick Shafto for useful discussions.
\end{ack}

\clearpage
\bibliographystyle{plainnat}
\bibliography{references}

\appendix
\raggedbottom

\section{Supplementary Results}
\zlabel{app:results}
\paragraph{Mini-Lean-agent with Opus 4.5.} We additionally run mini-Lean-agent once with Opus 4.5 on each problem of the Putnam 2025 benchmark. We run with a \$5 per-problem budget, 120-minute time limit, and 200 call limits (300 for A5/B6). The per-problem budget is slightly higher than we allow for Gemini 3.8 Flash, but is 10x smaller than the estimated per-problem cost of Numina-Lean-Agent. Unfortunately, we are unable to solve any of the 12 problems. Due to limited resources, and since we already obtain a strong result from more powerful models (Astra, Fable), we do not further pursue this direction. Note that a head-to-head comparison against Numina-Lean-Agent is not the goal of the study; instead, we showcase that mini-Lean-agent can show strong performance, when equipped with a powerful base model.

\paragraph{Quantitative and qualitative analysis of model outputs.}
\zcref{tab:rq1-tools,tab:rq1-usage} provide more details about the tool use and model usage in the \$2 budget runs for Gemini, excluding the extra runs for A5 and B6.

\zcref{fig:rq1-excerpts} shows excerpts of model outputs in the mini-Lean-agent and the AxProverBase harnesses. On A1, mini-Lean-agent submits a scratch proof, to which Lean returns a lemma that the agent decides to use in the final proof. On B3, a problem on modular arithmetic, AxProverBase makes 40 requests that invoke LeanSearch (the fuzzy search on a vector embeddings database), produces eight proof responses, and makes seven memory updates. After failing to close JSON strings in two proof responses, it leaves a \texttt{sorry} block in the final proof proposal, leaving the non-coprime case open.

\begin{table}[htbp]
  \centering
  \caption{\textbf{Number of tools requested through bash by mini-Lean-agent with Gemini and \$2 limit.} 
  The first column (bash) includes all requests, including malformed ones. ``Search'' refers commands like \texttt{rg} and \texttt{grep}. ``Lean'' refers to commands like \texttt{lake env lean} or \texttt{lake build}. Some tools may be counted in more than one column: e.g., a request can write a file with \texttt{cat} and invoke Lean. 
  }
  \zlabel{tab:rq1-tools}
  \PaperToolUseTable
\end{table}

\begin{table}[htbp]
  \centering
  \caption{\textbf{Model usage of mini-Lean-agent and AxProverBase with Gemini and \$2 limit.}
  Token counts sum across requests, including repeated context. Cache percentages use all input tokens as the denominator. Model calls include requests that return no bash action.}
  \zlabel{tab:rq1-usage}
  \PaperUsageTable
\end{table}

\begin{figure}[htbp]
  \small
  \textbf{mini-Lean-agent, A1.} The model writes: ``I'll test how to get an upper bound $K$ for a \texttt{Finset} $\mathbb{N}$ such that $t \subseteq \text{range } K$.'' It submits:
  \begin{lstlisting}[style=prompt,numbers=none,xleftmargin=0pt,framexleftmargin=0pt,literate={ℕ}{{$\mathbb{N}$}}1 {∃}{{$\exists$}}1 {⊆}{{$\subseteq$}}1]
cat <<'EOF' > Scratch.lean
import Mathlib

example (t : Finset ℕ) : ∃ K : ℕ, t ⊆ Finset.range K := by
  exact?
EOF
lake env lean Scratch.lean
\end{lstlisting}
  Lean returns \texttt{Try this: [apply] exact Finset.exists\_nat\_subset\_range t}.
  The accepted proof later uses this lemma.

  \medskip
  \textbf{AxProverBase, B3.} After two truncated proof responses, the memory model writes: ``Responses that exceed output token limits get truncated mid-string (\texttt{EOF while parsing a string}), leading to JSON parsing errors and failed builds.'' The final proof proposal still contains:
  \begin{lstlisting}[style=prompt,numbers=none,xleftmargin=0pt,framexleftmargin=0pt,literate={≡}{{$\equiv$}}1]
have h_modeq : 2025 ^ Nat.totient (n + 2) ≡
    15 ^ Nat.totient (n + 2) [MOD n + 2] := by
  apply Nat.ModEq.of_dvd
  sorry
\end{lstlisting}
  \caption{\textbf{Excerpts of Gemini's model output and tool feedback with \$2 limit.} 
  (Top) On A1, mini-Lean-agent sends a scratch-file query that contributes to a verified proof. (Bottom) On B3, AxProverBase submits an incomplete proof as its final proposal, after repeated retrievals and revisions. 
  }
  \zlabel{fig:rq1-excerpts}
\end{figure}

\clearpage
\paragraph{More detailed results.}
\zcref{tab:perproblem,tab:long-horizon} provide the per-problem statistics for the Gemini runs, respectively the \$2 limit run and the higher budget continuation. \zcref{tab:high-budget-per-problem} contains the per-problem statistics for the Astra and Fable. \zcref{tab:high-budget-checkpoints} shows the number of verified proofs at various thresholds, had the per-problem limit been lower for the Astra and Fable runs.

\begin{table}[htbp]
  \centering
  \caption{\textbf{Per-problem statistics with Gemini and \$2 limit.}
  This provides a detailed breakdown of the aggregate results in \zcref{tab:controlled}. ``mini'' refers to mini-Lean-agent and ``Ax'' refers to AxProverBase. The total charge does not include the charge for the higher budget run for A5/B6.
  Note that AxProverBase spent slightly over the \$2 cap on A1, but we mark it correct because no other call was made after the cap was reached.}
  \zlabel{tab:perproblem}
  \PaperPerProblemTable
\end{table}

\begin{table}[htbp]
  \centering
  \caption{\textbf{Per-problem statistics with Gemini and a higher limit on A5 and B6.}
  Each result includes the model charge for the first \$2 portion of the model output, where they continued from. If the final proof attempt was unsuccessful, the agent may stop before reaching any of the three limits.
  }
  \zlabel{tab:long-horizon}
  \PaperExtendedTable
\end{table}

\begin{table}[htbp]
  \centering
  \caption{\textbf{Per-problem statistics of mini-Lean-agent with Astra and Fable 5.1 at \$50 limit.}
  This provides a detailed breakdown of the aggregate results in \zcref{tab:high-budget}. Only Fable reaches its model charge limit on A5.}
  \zlabel{tab:high-budget-per-problem}
  \PaperHighBudgetPerProblemTable
\end{table}

\begin{table}[htbp]
  \centering
  \caption{\textbf{Per-problem statistics of mini-Lean-agent with Astra and Fable 5.1 at a lower budget limit.}
  Each column shows the number of verified problems out of 12 when the output is truncated at an even smaller per-problem limit.}
  \zlabel{tab:high-budget-checkpoints}
  \PaperHighBudgetCheckpointsTable
\end{table}

\paragraph{Total model charges.}
\zcref{tab:realized-charges} shows the total model charge for all reported experiments, including the twelve unsuccessful Opus 4.5 attempts.

\begin{table}[htbp]
  \centering
  \caption{\textbf{Total model charge.}
  Except for the second row, mini-Lean-agent is the harness used. Each row aggregates the model charge for exactly twelve problems. The first two rows include the higher budget continuations for A5 and B6.}
  \zlabel{tab:realized-charges}
  \PaperChargesTable
\end{table}

\section{Reproducibility}
\zlabel{app:reproducibility}
To make the proof counts and charges reproducible, we pin the benchmark, harnesses, dependencies, evaluator, and Lean/Mathlib \texttt{v4.28.0-rc1} and archive configurations, runs, usage records, and native LeanSearch queries and responses.
We route Gemini with high thinking to Google AI Studio and Opus with a 16,384-token thinking budget and 32,768-token output limit to Anthropic through OpenRouter, without provider fallback.
We route Astra (\texttt{gpt-6-astra}) and Fable (\texttt{claude-fable-5.1}) through OpenRouter to standard OpenAI and Anthropic, respectively, with \texttt{max} reasoning effort, a 128,000-token output limit, and no provider fallback.
Astra's recorded usage includes cache discounts, while Fable's contains no cached tokens.
We disable networking in proof and evaluator containers (the host mediates AxProverBase's LeanSearch requests).
The evaluator preserves the original source prefix, permits only plain theorem declarations, and rejects incomplete proofs, extra axioms, and metaprogramming.
Fresh independent Comparator instances verify each target, allowing only \texttt{propext}, \texttt{Quot.sound}, and \texttt{Classical.choice}, with a 600-second limit per snapshot~\citep{noauthor_leanprovercomparator_2026}.

\clearpage
\section{Prompts}
\zlabel{app:prompts}
The prompts specify the task and submission rules that the evaluator enforces.
We give mini-Lean-agent and AxProverBase the same task instruction:
\begin{quote}
  \small
  Complete the proof of the designated theorem in \texttt{Target.lean}.
  The registered project uses Lean 4.28.0-rc1 and Mathlib v4.28.0-rc1.
  Preserve the supplied statement, definitions, imports, and project configuration.
  Keep the supplied source prefix before the target theorem byte-identical.
  You may add proved helper theorems after that prefix, using plain \texttt{theorem} commands (not \texttt{lemma}, declaration attributes, or modifiers), and create scratch files while working.
  Submit \texttt{Target.lean} with no \texttt{sorry}, \texttt{admit}, additional axioms, or metaprogramming commands.
  The final artifact is checked independently in the original pinned environment.
  Use only the supplied tools; external models and open-web browsing are forbidden.
\end{quote}

For mini-Lean-agent, we use the following system and initial user templates.

\paragraph{System message template.}
\leavevmode\par
\begin{lstlisting}[style=prompt]
You are a helpful Lean 4 theorem-proving assistant that can interact with a computer shell.
\end{lstlisting}

\paragraph{Initial user message template.}
\leavevmode\par
\begin{lstlisting}[style=prompt]
Please solve this Lean theorem-proving task: {{task}}

You can execute bash commands and edit files to complete the proof.
Your working directory is {{lean_project}}.
The fixed target file is {{lean_target}}.
The protected declaration is {{lean_declaration}}. Its statement must match the trusted challenge when your proof is verified.
Use the definitions, lemmas, and tactics available through the target's imports.

## Recommended Workflow

Work step-by-step so that you can use Lean's feedback to improve the proof.

1. Read the task, fixed target, and its imports to understand the informal and formal statements
2. Identify relevant definitions, lemmas, and proof strategies. Search the project and dependencies, and use small scratch Lean files to check declaration names instead of guessing
3. Edit {{lean_target}} with a complete proof
4. Run `{{lean_check_command}}` after each proof attempt
5. Read every error or warning, revise the proof, and check it again
6. Once the check succeeds, submit your proof and finish your work by issuing the following command: `echo COMPLETE_TASK_AND_SUBMIT_FINAL_OUTPUT`.
   Do not combine it with any other command. <important>After this command, you cannot continue working on this task.</important>

## Correctness Rules

- Do not change or weaken the theorem statement, or replace it with an easier statement.
- Do not use `sorry`, `admit`, or `sorryAx`, introduce new axioms, or otherwise bypass the proof.
- Leave the complete checked proof in {{lean_target}}.
- Fix every compiler error. Treat any warning that a declaration uses `sorry` as a failed proof.

## Command Execution Rules

You are operating in an environment where

1. You issue at least one command
2. The system executes the command(s) in a subshell
3. You see the result(s)
4. You write your next command(s)

Each response should include:

1. **Reasoning text** where you briefly explain your analysis and next step
2. At least one tool call with your command

**CRITICAL REQUIREMENTS:**

- Your response SHOULD include reasoning text explaining what you're doing
- Your response MUST include AT LEAST ONE bash tool call
- Directory or environment variable changes are not persistent. Every action is executed in a new subshell.
- Combine dependent operations with `&&`, or repeat any required `cd` and environment setup.
- Use non-interactive commands.
- Submit your proof with the stand-alone command `echo COMPLETE_TASK_AND_SUBMIT_FINAL_OUTPUT` after the check succeeds.
  The harness verifies the fixed target and its protected statement after the agent exits.
  <important>After this command, you cannot continue working on this task.</important>

Example of a CORRECT response:
<example_response>
I need to inspect the target and its imports before choosing a proof strategy. Let me locate the Lean source files.

[Makes bash tool call with {"command": "rg --files -g '*.lean'"} as arguments]
</example_response>

<system_information>
{{system}} {{release}} {{version}} {{machine}}
</system_information>

## Useful command examples

Replace example paths and declaration names with those relevant to your task.

### Create a scratch Lean file:

Use the target's imports when exploring its libraries.

```bash
cat <<'EOF' > Scratch.lean
import Lean
#check And.intro
#check And.left
EOF
```

### Find relevant definitions and lemmas:

```bash
rg -n 'declaration_name' . --glob '*.lean'
```

Dependency sources, when present, can be searched directly:

```bash
rg -n 'declaration_name' .lake/packages --glob '*.lean'
```

### View file content:

```bash
nl -ba path/to/Target.lean | sed -n '1,160p'
```

### Check the proof:

```bash
{{lean_check_command}}
```
\end{lstlisting}

\end{document}